\documentclass[letterpaper, 10 pt, conference]{ieeeconf}  % Comment this line out if you need a4paper
\usepackage{amsmath}
\usepackage{algpseudocode} 
\usepackage{algorithm}
\usepackage{booktabs}
\usepackage{siunitx}

\usepackage[switch]{lineno}
\usepackage{xcolor}

\newcommand\scomment[1]{}

\IEEEoverridecommandlockouts                              % This command is only needed if 
\usepackage{graphics} % for pdf, bitmapped graphics files
\usepackage{epsfig} % for postscript graphics files
\usepackage{mathptmx} % assumes new font selection scheme installed
\usepackage{times} % assumes new font selection scheme installed
\usepackage{amsmath} % assumes amsmath package installed
\usepackage{amssymb}  % assumes amsmath package installed

\title{\LARGE \bf
Learning Attentive Neural Processes for Planning with Pushing Actions
}

\author{Atharv Jain$^{1}$, Seiji Shaw$^{1}$, Nicholas Roy$^{1}$% <-this % stops a space
\thanks{$^{1}$Computer Science and Artificial Intelligence Laboratory, Massachusetts Institute of Technology 
       }%
}
\begin{document}

\maketitle
\thispagestyle{empty}
\pagestyle{empty}

%%%%%%%%%%%%%%%%%%%%%%%%%%%%%%%%%%%%%%%%%%%%%%%%%%%%%%%%%%%%%%%%%%%%%%%%%%%%%%%%
\begin{abstract}
% Our goal is to enable robots to plan sequences of tabletop actions to push a block with unknown physical properties to a desired goal pose on the table.
% We approach this problem by learning the constituent models of a Partially-Observable Markov Decision Process (POMDP), where the robot can observe the outcome of a push, but the physical properties of the block that govern the dynamics remain unknown.
% The pushing problem is a difficult POMDP to solve due to the challenge of state estimation. 
% The physical properties have a nonlinear relationship with the outcomes, requiring computationally expensive methods, such as particle filters, to represent beliefs. 
% Leveraging the Attentive Neural Process architecture, we propose to replace the particle filter with a neural network that learns the inference computation over the physical properties given a history of actions. 
% This Neural Process is integrated into planning as the Neural Process Tree with Double Progressive Widening (NPT-DPW). 
% Simulation results indicate that NPT-DPW generates more effective plans faster than traditional particle filter methods, even in complex pushing scenarios.

Our goal is to enable robots to plan sequences of tabletop actions to push a block with unknown physical properties to a desired goal pose. 
We approach this problem by learning the constituent models of a Partially-Observable Markov Decision Process (POMDP), where the robot can observe the outcome of a push, but the physical properties of the block that govern the dynamics remain unknown. 
A common solution approach is to train an observation model in a supervised fashion, and do inference with a general inference technique such as particle filters. 
However, supervised training requires knowledge of the relevant physical properties that determine the problem dynamics, which we do not assume to be known.
Planning also requires simulating many belief updates, which becomes expensive when using particle filters to represent the belief. 
We propose to learn an Attentive Neural Process that computes the belief over a learned latent representation of the relevant physical properties given a history of actions. 
To address the pushing planning problem, we integrate a trained Neural Process with a double-progressive widening sampling strategy. 
Simulation results indicate that Neural Process Tree with Double Progressive Widening (NPT-DPW) generates better-performing plans faster than traditional particle-filter methods that use a supervised-trained observation model, even in complex pushing scenarios.

\end{abstract}

%%%%%%%%%%%%%%%%%%%%%%%%%%%%%%%%%%%%%%%%%%%%%%%%%%%%%%%%%%%%%%%%%%%%%%%%%%%%%%%%
\section{Introduction}

In scenarios where robots operate in unstructured, dynamic environments, planning under uncertainty is paramount. Some relevant properties for manipulation, such as object geometry, can be easily observed through sensors. Many other properties, such as inertial properties and friction, require interaction with the object \cite{bohg_interactive_2017}. The robot may not even know what the relevant physical properties are to determine action outcomes. One way robots can learn these inertial properties is to iteratively interact with objects and observe their behavior.  

We focus on an instance of this problem where a robot must compose a sequence of pushes to guide a block from a starting configuration to a goal configuration, without prior knowledge of the center of mass. Our goal is to develop a technique capable of inferring the true center of mass while simultaneously predicting outcomes of pushing actions. 

We model our planning challenge as a Partially Observable Markov Decision Process (POMDP), where unobservable physical properties remain constant over time \cite{doshi-velez_hidden_2013}. Many POMDP solutions rely on Bayesian state estimation techniques (e.g., Kalman or particle filters) to update their beliefs as new observations become available. However, the linear assumptions required for Kalman filters do not hold in this scenario, since varying the center of mass can nonlinearly influence rotational outcomes. More general methods, such as particle filters, must be used, despite their high computational cost \cite{zhang_application_2012a}. 

To leverage these particle-based belief representations to solve our planning problem, a common method is to use \textit{Particle Filter Trees with Double Progressive Widening} (PFT-DPW), a Monte Carlo tree search algorithm that performs a randomized search to identify the plan yielding the highest expected reward. In POMDP planning, it is crucial to balance exploration and exploitation \cite{kaelbling_planning_1998}. PFT-DPW and similar Monte Carlo tree search methods manage this trade-off effectively by systematically exploring uncertain aspects of the system while exploiting known information to achieve high rewards \cite{sunberg_online_2018}. 

However, PFT-DPW also has significant limitations due to the use of a particle filter. First, training an observation model used to update the particle filter requires knowing what the relevant physical properties of the object for the task dynamics are. Second, the runtime cost of updating a particle belief scales linearly with the number of particles when searching for long-horizon plans. %Since belief updates occur repeatedly as a subroutine within the planning algorithm, these costs accumulate quickly. 
% \sas{footnote?} 
% Second, PFT-DPW is limited in its ability to handle more complex problems: it remains susceptible to the \textit{curse of dimensionality}, where the required number of particles grows exponentially with the dimension of the underlying distribution, making it difficult to scale to higher-dimensional tasks. \aj{yeah so this part is a bit deceptive but not technically a lie... we don't actually test it in high dim scenarios but i did include it as a weakness. honestly okay to keep but maybe need to look at.}

To overcome these problems, we propose a model called the Pushing Neural Process (PNP), which jointly trains an inference network that estimates physical parameters, and a prediction network that learns to output a distribution of push outcomes based on physical properties. The inference network and prediction network are set up in an encoder-decoder architecture, where a representation of the physical properties is learned in the intermediate latent space. Notably, the PNP does not require any information about which physical properties are relevant to the problem, but instead encodes any relevant information about the representation into a multivariate gaussian outputted by the encoder. 

To compare the performance of both models, we derive a new planner that integrates the PNP, called NPT-DPW (Neural Process Tree with Double Progressive Widening). NPT-DPW replaces the particle filter used to maintain belief in PFT-DPW with a neural network that consumes the history of a given search and learns a representation for the object's unknown physical properties. Our results show that, on average, NPT-DPW reaches up to twice as close to the goal location as PFT-DPW and avoids catastrophic failures (such as pushing into prohibited zones) far more often. We additionally demonstrate that the PNP's latent space embeds a representation of the true COM by showing that the model becomes more accurate with greater amounts of previous pushes.

\begin{figure*}[htbp]
    \centering
    \includegraphics[width=\textwidth]{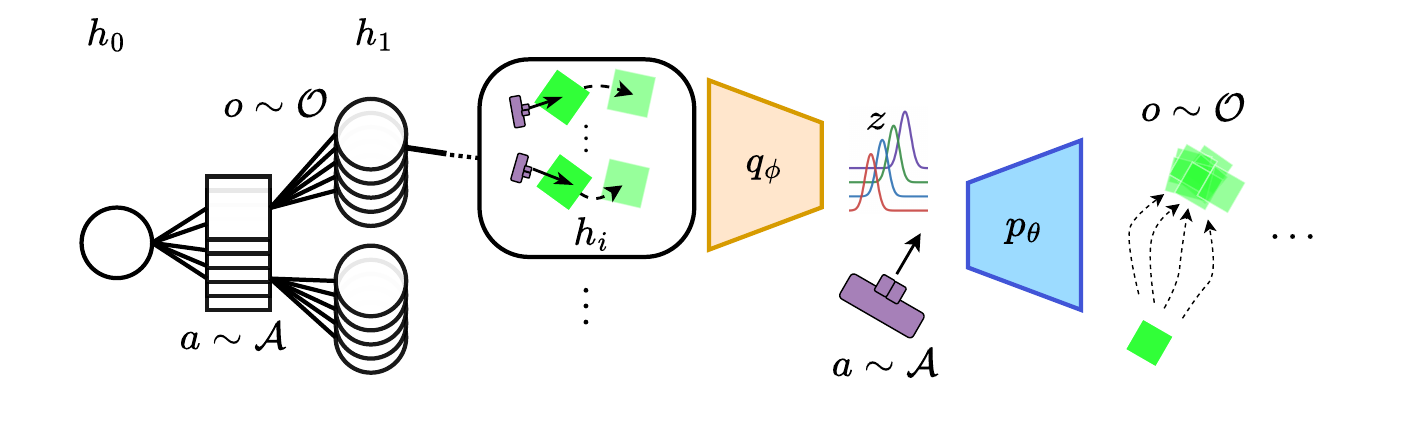}
    \caption{The Push Neural Process architecture and its use as a subroutine within NPT-DPW. On the right of the image, the search tree over actions and outcomes is pictured, where at each step an action is selected. We then pass in the history of outcomes ($h$) that we have already observed into the encoder ($q_\phi$) of the model to obtain the distribution over the latent parameters, which we sample to get ($z$). We then pass $z$ to the decoder with our new action $a$ to get a distribution over outcomes. We sample from that distribution as an outcome ($o$) for the action $a$. We then add that outcome to the potential outcomes in the search tree.}
    \label{fig:wide-figure}
\end{figure*}
\section{Problem Setting}

\begin{figure}
    \centering
    \includegraphics[width=0.8\linewidth]{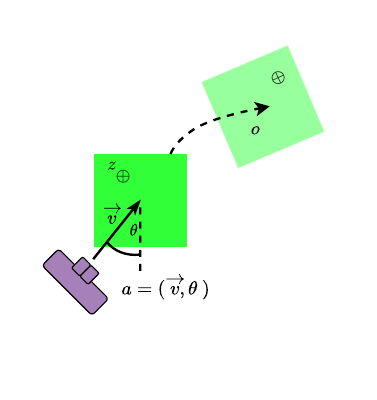}
    \vspace{-1.0cm}
    \caption{A depiction of a push action $a \in \mathcal A$, parameterized by push angle $\theta \in [0, 2\pi)$ and velocity $\overrightarrow v \in \mathbb R^2$. The center of mass $z$ (denoted as $\oplus$) is not directly observed.}
    \label{fig:push}
\end{figure}

For the objects we aim to push, we categorize their properties into two distinct sets: observable properties, denoted by $x \in \mathcal{X} \subset \mathbb R^n$, and latent physical properties, denoted by \(z \in \mathcal{Z} \subset \mathbb R^m \), where $m$ is decided empirically based on what results in the strongest model. The observable properties \(x\) include features such as the block's resting orientation, while \(z\) captures physical properties such as center of mass. These latent properties are unknown to the robot \emph{a priori} but play a critical role in determining the outcome of a push.
We define a push in terms of an action $a \in \mathcal{A}$, characterized by an approach angle and a push velocity, where the angle is in the range $[0,2\pi)$. 
Upon completing a push, the robot observes the resulting pose of the block $o \in SE(2)$. A schematic of the push action is depicted in Figure \ref{fig:push}.

Within our Partially Observable Markov Decision Process (POMDP) framework, the \emph{transition function} $T$ models the object's dynamics. It maps the current state—which encompasses both the object's observable pose and its latent physical properties—and an action  to a probability distribution over future object poses. Importantly, the object's pose remains observable to the robot at all times, while the latent properties remain constant throughout the interaction.

Consequently, our \emph{observation function} $O$ simply reflects the observable portion of the state provided by the transition function $T$. The \emph{reward function} $R$ penalizes the robot proportionally to its distance from the target goal, offering a substantial reward upon successfully reaching the goal (see Section \ref{sec:exp-scenario}, Equation \ref{eqn:reward} for explicit details). A discount factor $\gamma$ further encourages shorter plans aimed at efficiently reaching the goal.

Collectively, we formulate our pushing task as a POMDP defined by the tuple 
\[
(\mathcal{S}, \mathcal{A}, \Omega, T, O, R, \gamma)
\]
where $\mathcal{S} = \mathcal{X}\times \mathcal{Z}$ represents all possible states and $\omega\in SE(2)$ represents all possible observations. 

\section{Methods}\label{sec:methods}

\subsection{Pushing Neural Process}\label{sec:pnp}
To solve the POMDP, we require probabilistic models that describe transitions in the partially observable state of the system and the observations made by the robot. 
We use an Attentive Neural Process, which we call a \textit{Pushing Neural Process}, to jointly model push state estimation and prediction \cite{kim_attentive_2018}. This model consists of two primary components: an \textit{encoder} that infers latent representations from historical data, and a \textit{decoder} that predicts future push outcomes given these representations.

We denote a sequence of pushes and their outcomes as the dataset:
\[
D_t = \{(a_1, o_1), \ldots, (a_t, o_t)\}.
\]
The encoder consumes $D_t$ (of arbitrary length) along with the observable properties $x$ (object geometry) and infers a probability distribution over the latent parameters $z$. Although the latent space is uninterpretable, we hypothesize that the model will implicitly learn what useful information can be learned from pushing.  To this end, we train a neural network, parameterized by $\phi$, to produce an approximate posterior $q_\phi(z|x, D_t)$ that closely matches the true posterior $p(z|x, D_t)$.

Given the latent representation from the encoder, the decoder—parameterized by $\theta$—then predicts push outcomes. Mathematically, prediction of a new outcome $o$ given action $a$, object properties $x$, and history $D_t$ can be expressed using the law of total probability: 
\[
p(o|a,x,D_t)=\int p_\theta(o|a,x,z)p(z|x, D_t)\,dz.
\]

\begin{algorithm}
\caption{NPT-DPW}\label{alg:cap}
\begin{algorithmic}
\Procedure{SimulateAction}{$h,a$}   
    \State $b\gets $\Call{PNPEncoder}{h} \Comment{PNP inference step.}
    \State $p \gets$ \Call{PNPDecoder}{b, a} \Comment{PNP prediction step.}
    \State $o \gets p.sample()$ 
    \State $r \gets \text{Reward}(o.x,o.y)$ \Comment{As defined in Equation~\ref{eqn:reward}}
    \State \Return $o,r$
\EndProcedure

\Procedure{Simulate}{$h,d$}
\If{$d=0$}
    \State \Return 0
\EndIf
\State $a \gets$ \Call{ActionProgWiden}{h}
\If{$|C(ha)|\leq N(ha)^{\alpha_0}$}
    \State $o, r \gets$ \Call{SimulateAction}{h,a}
    \State $C(ha)\gets C(ha)\cup\{o\}$
    \State $total\gets r+\gamma\cdot$\Call{Rollout}{$hao, d-1$}
\Else
    \State $o,r \gets$ sample from $C(ha)$ 
    \State $total\gets r+\gamma\cdot$\Call{Simulate}{$hao, d-1$} 
\EndIf
\State \Return $total$
\EndProcedure
\end{algorithmic}
\end{algorithm}

Since a single sample from $p(z|x, D_t)$ is an unbiased estimator of the distribution, we sample a single $z \sim q_\phi(z|x, D_t)$. The decoder consumes $z$, the observable object properties, and the new push to predict a distribution over the observation (block displacement) $p_\theta(o|a,x,z)$.

During training, we maintain a collection of pushes for each object in a set of objects with varying physical properties. We provide the partial dataset $D_t$ as input to the inference network (encoder), and apply the prediction network (decoder) to the collection of \textit{all} pushes $D_T$ for the given object:
\[
D_T = D_t \cup \{(a_{t+1}, o_{t+1}), \ldots, (a_{T}, o_T)\}.
\]
We optimize the Evidence Lower Bound (ELBO)
\[
\mathbb{E}\Bigl[\log p_\theta(o|x,a,z) \;-\; D_{KL}\bigl(q_\phi(z|D_T)\;\|\;q_\phi(z|D_t)\bigr)\Bigr],
\]
which promotes accurate push outcome predictions while reducing the discrepancy between the full and partial posterior distributions. The cross-entropy term increases the likelihood of accurately predicting outcomes over time, and the KL term mitigates overfitting.

\subsection{NPT-DPW}
Our planner is an extension of PFT-DPW introduced by Sunberg and Kochenderfer \cite{sunberg_online_2018}. 
In PFT-DPW, a particle filter is used to represent the set of hypotheses over the partially observable state of the problem, and the belief is updated as new push outcomes are observed.
The PF is used as a belief state in Monte Carlo tree search, which branches at new actions and observations. 
Since both the action and observation spaces are continuous, double progressive widening is used to limit the number of actions and observations sampled at each tree node to ensure reliable reward estimates.

Instead of using a particle filter, our planner, NPT-DPW, uses the Pushing Neural Process to infer the underlying physical parameters and predict outcomes of future actions.
To expand the tree, we pass the history of actions and outcomes through the encoder to infer the physical parameters. This history of actions implicitly represents a belief of the true properties, as the history of actions can be passed into the encoder to generate a distribution of the true physical properties. 

Actions are sampled as they normally would be in PFT-DPW.
Observations are sampled from the distribution computed by the decoder from the inferred parameters and the push under consideration. 
Thus, every step of the tree search will have runtime complexity of \(O(h^2)\), where \(h\) is the size of the history (due to the quadratic complexity of self-attention), rather than \(O(n)\), where \(n\) is the number of particles used in PFT-DPW. 
We hypothesize a significant improvement in plan quality within a fixed computational budget, since \(h \ll n\) in practice. As stated previously, an additional advantage of this method is that it works in general scenarios where the relevant properties are not known \textit{a priori} to the robot. 

The NPT-DPW algorithm is described in Algorithm~\ref{alg:cap}; any functions not described in this block are the same as those specified for PFT-DPW in Sunberg and Kochenderfer \cite{sunberg_online_2018}. Furthermore, a diagram of the Neural Process and its integration into the planner can be seen in Figure~\ref{fig:wide-figure}.

\begin{figure}
    \centering
    \includegraphics[width=1\linewidth]{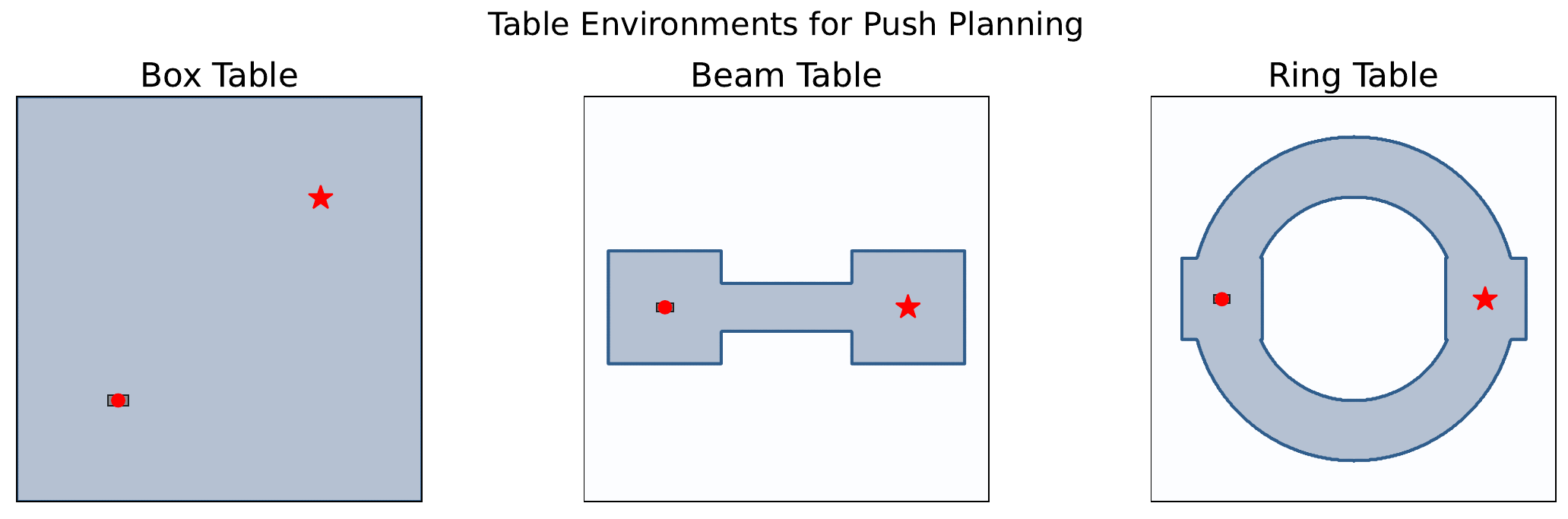}
    \caption{The pushing surfaces used in our experiments. The rectangle denotes the starting position, and the star the goal.}
    \label{fig:enter-label}
\end{figure}

\section{Experiments}
\begin{figure*}[htbp]
    \centering
    \includegraphics[width=1\linewidth]{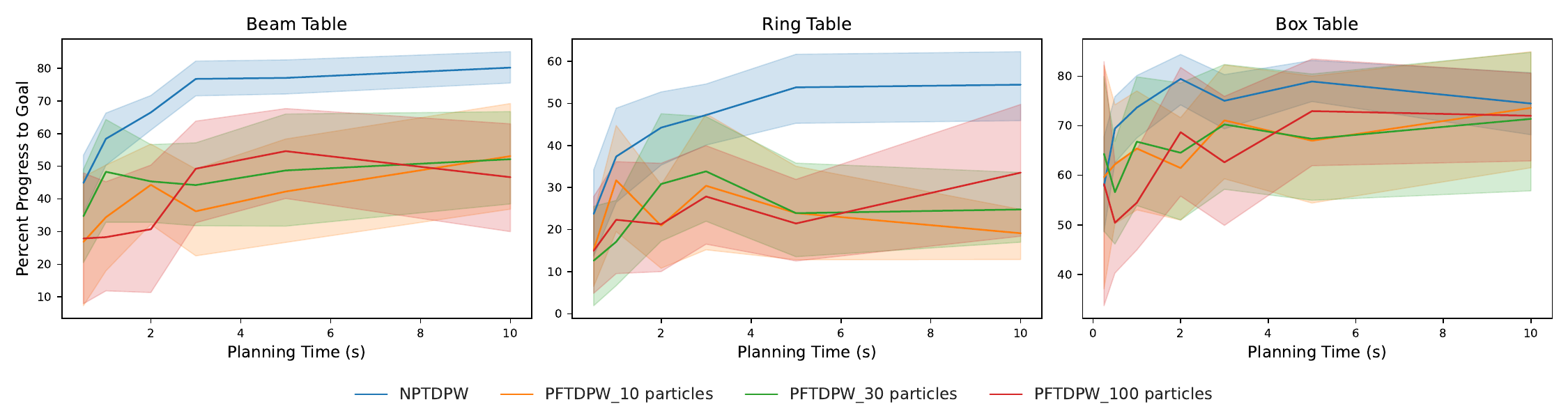}
    \caption{This figure demonstrates the effectiveness of the NPT-DPW versus PFT-DPW (with 10/30/100 particles). Additionally, we conducted trials with 0.5,1,2,3,5, and 10 seconds of time to plan per step. The results plot the average percentage traveled towards the goal as a function of planning time allocated per action. For example, if the block started 3 meters away, and was pushed to being 1 meter way, that would be considered 66\% progress. Fallen blocks are marked with the percentage distance corresponding to when they fell off. Therefore, the percentage from the goal varies a lot, since times the block will fall off almost immediately will be marked with an approximately 0\% success. }
    \label{fig:results}
\end{figure*}
In our experiments, we seek to evaluate the performance of NPT-DPW relative to the baseline method PFT-DPW, given a fixed computational runtime budget.
The problem scenarios are designed to evaluate whether an MCTS-DPW planning scheme can leverage the efficient inference updates of the Neural Process for a deeper tree search, yielding better plans.

\subsection{Scenarios}\label{sec:exp-scenario}

The experiments will be run using three tables with geometry designed to vary the hardness of the pushing problem (Figure \ref{fig:enter-label}). 
Each of these tables defines a different planning problem. The first is a blank-surface table designed to check whether the planner can produce plans that drive the block to the goal. 
In the second and third scenarios, the movement of the block must be constrained along certain paths to prevent it from falling off the pushing surface. The second table requires pushing in a straight line, while the third requires a push trajectory along a curve. 

Solving these instances requires the planner to accurately infer the object’s physical properties.
For the second and third tables, an intelligent planner must sequence information-gathering actions on a wider area before pushing the block along the narrow sections of these tables. 
       
\begin{figure}
    \centering
    \includegraphics[width=1\linewidth]{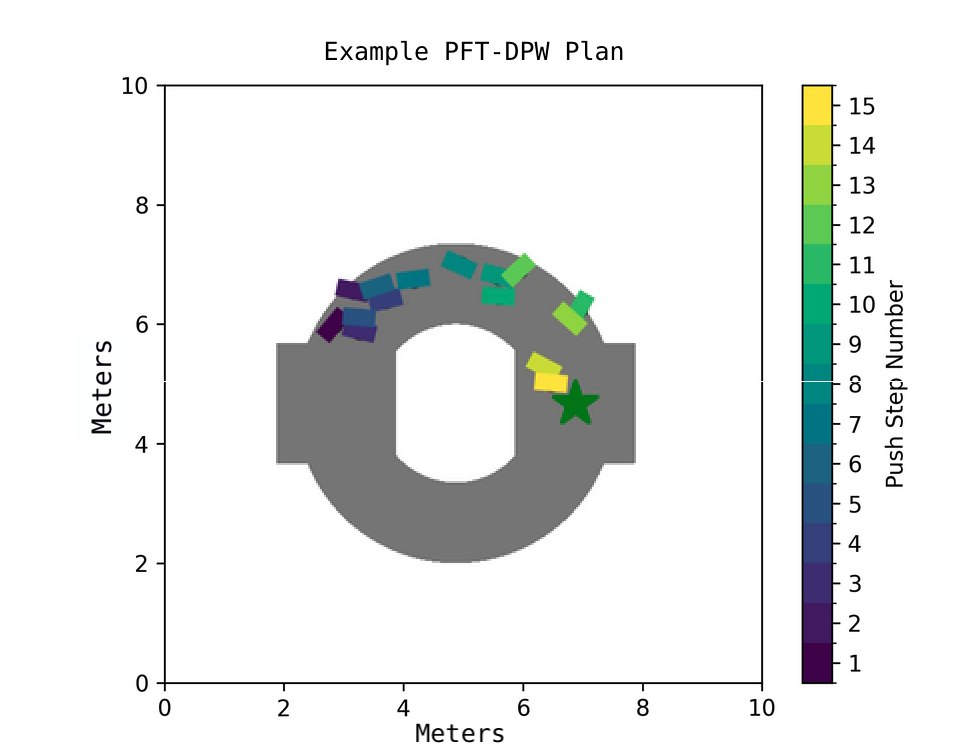}
    \caption{This is an example of a plan created by NPT-DPW and run by the simulation on the "ring" map. Important features that show signs of intelligent planning are the pushes in the beginning in the corner, signifying learning more about the block before trying to move towards the goal. Note the figure has been re-scaled slightly from the real experiments to be more visually clear.}
    \label{fig:example-plan}
\end{figure}

After each push, the robot obtains a reward for the block’s location \((x, y)\) relative to the goal \((g_x, g_y)\). We define the reward function
\begin{align}
\text{Reward}(x, y) = &-10\sqrt{(x - g_x)^2 + (y - g_y)^2} \nonumber\\
& - 100 \cdot \texttt{fallen}(x, y) \nonumber \\
& + 10000 \cdot \texttt{inRange}(x, y) \label{eqn:reward}
\end{align}
\noindent $\texttt{fallen}$ is a function that returns 1 if the block is off the table, and 0 otherwise. $\texttt{inRange}$ is a function that is $1$ if the block is within 0.2 meters of the goal, and $0$ otherwise. We set the discount factor $\gamma = 0.8$ to incentivize the planner to prioritize not push the block off the table. An example of a plan generated by NPT-DPW is depicted in Fig. \ref{fig:example-plan}. While the metric used as the model to plan will be the reward function, we do not evaluate performance based on that, but instead use percent progress towards the goal, as can be seen later in the results section. 

We assess performance by fixing the computation time $t$ allowed at each step of the plan. Then for each $t$ in $[0.5,1,2,3,5,10]$ seconds, we ran 15 trials on each map for each planner, where we recorded the distance of the object from the goal after 30 steps. If the object fell off the table, or reached the goal early, we stopped then. We sampled a new center of mass for the block in each trial. For each tree search, we looked for the best action for depth equaling 3, and we only stored the first 10 actions for NPT-DPW, since empirically that was enough to gain a strong belief over the block. 

We also varied the number of particles in PFT-DPW, initializing it with $n=10,30$, or $100$, to evaluate NPT-DPW against a range of particle filter–based planners and assess the impact of particle count on performance. Additionally, while maintaining the particle filter, if the weight associated with a particle became smaller than $10^{-9}$, then we turn off that particle, which should encourage the particle filter to become faster as plans become deeper, since low weighted particles will no longer be counted. 

These simulations are run in the PyBullet simulator. 
We implement an end-effector PD-controller to ensure the gripper's end-effector motion velocity is consistent throughout the motion. 
To sample actions for progressive widening, we uniformly sample an angle from $[0, 2\pi]$ in world coordinates and push at the velocity of $0.10$ m/sec. 
Each of the blocks is a rectangular prism with a fixed geometry of 2.5 cm x 2.5 cm x 1.5 cm. 
The mass of the block is set constant at 0.2 kg. 
The center of mass of the block is sampled uniformly within the convex hull of the object's geometry.
% In our experiments, we define a "push" to be trajectory followed by the robot gripper moving at a specified heading and velocity for three seconds. 
% The robot will not readjust the gripper if the block slips or rotates over the course of the push trajectory.\footnote{If the block travel's outside the robot's workspace, we move the robot to be sufficiently close object to execute the next push action.}

Our neural network architecture is based on Attentive Neural Processes \cite{kim_attentive_2018}. 
Each push is first encoded by an MLP and processed through multiple layers of self-attention. After that, the outputs of the self-attention layers are aggregated into outputs consisting of a mean and covariance. These represent the mean and diagonal covariance of a multivariate Gaussian over $z$, representing the distribution over latent physical properties. Note that for our problem, the only latent physical properties are the $x$ and $y$ coordinates of the center of mass, but we still set $z\in \mathbb{R}^5$ since it empirically led to improved performance. 
The decoder is structured as multiple linear layers that consume a sample from the latent space and outputs a distribution over possible outcome poses of the block after the push is finished.

\sisetup{
  round-mode          = places,
  round-precision     = 0,
  table-number-alignment = center,
}

\begin{table}[htbp]
  \centering
  \caption{Number of Simulations Done with Different Amounts of Planning Time, Averaged Over 10 Plans}
  \label{tab:simulate-calls}
  \begin{tabular}{@{} l *{3}{S[table-format=4.1]} @{}}
    \toprule
    Planner                       & {1\,s} & {5\,s} & {10\,s} \\
    \midrule
    NPTDPW                      &  325.3   & 2291   & 4807   \\
    PFTDPW with 10 Particles      &  157.6   &  806.6   & 1910   \\
    PFTDPW with 30 Particles      &   62.80   &  646.0   & 1410   \\
    PFTDPW with 100 Particles     &   30.20   &  322.2   &  977.1   \\
    \bottomrule
  \end{tabular}
\end{table}

\subsection{Results}

Our results, shown in Figure~\ref{fig:results}, highlight the effectiveness of using a learned latent space for efficient planning. NPT-DPW consistently outperforms PFT-DPW across multiple scenarios when allocated at least one second per action. This performance gain suggests that the latent space learned by the Attentive Neural Process provides a compact yet expressive understanding of the block's unknown physical properties, which NPT-DPW was able to leverage to achieve greater success planning. 
\begin{figure}
    \centering
    \includegraphics[width=1\linewidth]{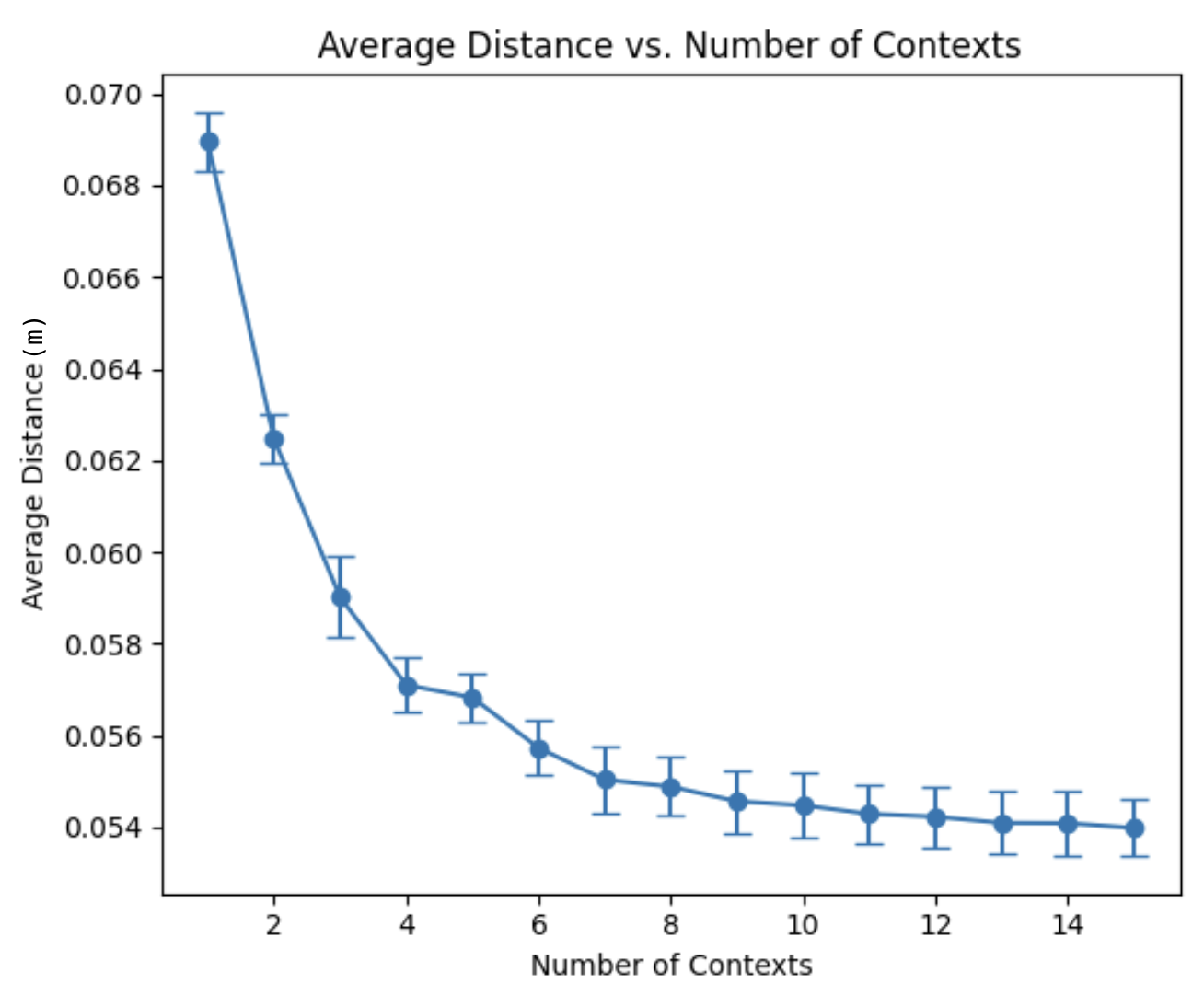}
    \caption{ This plot shows how the average error of the predicted location of the block from the Push Neural Process to the true result of the location of that push decreases as the model receives more past interactions, or context pushes. The error bars represent an interval of one standard error around the mean distance. }
    \label{fig:latentincrease}
\end{figure}

Figure \ref{fig:latentincrease} demonstrates that knowing the result of previous interactions with a block (called "context pushes") enables the robot to make more accurate pushes, since the robot is able to process those previous outcomes in the encoder. Because of this, we can infer that the PNP's latent space is able to learn a representation of the block's center of mass.

In addition, maintaining belief through the latent space reduces the computational overhead of having runtime proportional to the number of particles of explicit particle representations, which is what we hypothesize allows NPT-DPW to more rapidly explore the space of potential action sequences than PFT-DPW. In specific, in Table \ref{tab:simulate-calls}, we have shown that the number of planning iterations the PNP can process is significantly higher than the particle filter even when the number of particles is 10. 

While all these results are extremely promising, there are a few caveats. In scenarios where very limited planning time (e.g., half a second per action) was available, both models struggled similarly, underscoring the necessity for adequate computation time to fully leverage the benefits of learned latent spaces. NPT-DPW, however, starts out performing PFT-DPW when the amount of computation becomes greater than a second. 

 Another interesting result of the data is the lack of difference in performance between the varied number of particles in PFT-DPW. We hypothesize this is due to the nature of the pushing task, where subtle differences in the center of mass have limited impact because push outcomes primarily depend on the relative positions of the center of mass and the point of contact. As a result, a small number of particles might already capture most of the informative uncertainty, while increasing particle counts does not significantly enhance prediction quality but rather increases computational load, limiting the planner's capacity for deeper exploration.

Our findings reinforce the potential of latent spaces in the Attentive Neural Process to maintain a useful learned representation of the unobservable properties of an object. Additionally, we showed that using the Attentive Neural Process in a planner leads to significantly improved performance compared to PFT-DPW. These results indicate promising directions for future exploration into richer representation learning, aiming to enhance generalization and robustness in robotic planning tasks.

\section{Related Work}

Our pushing POMDP is an instance of a Hidden-Parameter POMDP \cite{doshi-velez_hidden_2013}, where the unobserved state variables remain static but influence the transition dynamics of the planning problem.
Several previous works have introduced Bayesian models for rapid estimation and adaptation of the unobserved state in this class of problems. 
The difference between these methods largely depends on the representation of uncertainty.
Killian et al. \cite{killian_robust_2017} use Bayesian neural nets; others use Gaussian processes \cite{doshi-velez_hidden_2013,saemundsson_meta_2018}.
Ensembles of neural networks are a valid choice as well \cite{belkhale_model-based_2021}.

Other work has focused on learning object dynamics with unobservable object properties outside the POMDP framework. 
Many authors develop ways to estimate unobserved properties when the dynamics are known \cite{battaglia_simulation_2013, wu_galileo_2015, wu_physics_2016, zhang_application_2012}. 
Other works, like our method, do not assume the dynamics are known \cite{lu_extracting_2019, veerapaneni_entity_2019, xu_densephysnet_2019, zheng_unsupervised_2018}.
However, these works assume they have access to a dataset of interactions from a new instance of the system to estimate the unobserved variables. 

Estimating unobserved properties through interaction is a common strategy in many manipulation tasks \cite{bohg_interactive_2017}. 
These strategies are applied to adapt grasping classifiers
\cite{noseworthy_amortized_2024,danielczuk_exploratory_2021,fu_legs_2022,zhao_center_2018,cehajic_estimating_2017}, disentangle object geometry without visual input \cite{herzog_learning_2014,dragiev_uncertainty_2013}, and obtain object articulation constraints \cite{gadre_act_2021,martin-martin_coupled_2022}.
Bauza and Rodriguez \cite{bauzaGPSUMGaussian2020} develop an alternate method to compute uncertainty of push outcomes based on uncertainty of physical parameters of known relevance.
Many of these methods tackle information gathering by developing problem-specific solutions \cite{zhao_center_2018,cehajic_estimating_2017}, or relying on bandit-style exploration approaches \cite{danielczuk_exploratory_2021,fu_legs_2022} or information-gain heuristics \cite{noseworthy_amortized_2024,dragiev_uncertainty_2013} for short, single-step estimation tasks.
Our work leverages POMDP solvers to tackle the information-gathering problem when multiple actions must be composed together.

\section{Conclusion}

We present: the Pushing Neural Process, an Attentive Neural Process architecture to learn push outcome prediction of a novel object instance, and NPT-DPW, a new planner that integrates the PNP to produce good plans under reasonable computational runtime budgets. Additionally, we demonstrated a method of learning a representation of an object's properties, and how we can leverage that representation to maintain belief with a general use case planner. 

Our initial experiments demonstrate promising capabilities; however, we acknowledge certain limitations that we aim to address in subsequent research. For instance, the PNP exhibited underfitting, which we attribute to the complexity and inherent instability of the simulation environment used to generate the training data. To address this, we hope to apply the NPT-DPW to a broader spectrum of tasks, including scenarios with simpler dynamics and transition models to yield clearer insights. Secondly, we observed occasional erratic behavior in generated plans, indicating opportunities for further model improvement. Addressing these issues may improve planning performance in both PFT-DPW and NPT-DPW. 
Finally, we hope to deploy the PNP on a physical robot platform to comprehensively validate our approach and demonstrate its applicability.

\section*{ACKNOWLEDGMENTS}

We thank Michael Noseworthy and Ben Zandonati for their helpful feedback and support.

%%%%%%%%%%%%%%%%%%%%%%%%%%%%%%%%%%%%%%%%%%%%%%%%%%%%%%%%%%%%%%%%%%%%%%%%%%%%%%%%

% \begin{thebibliography}{99}

% \end{thebibliography}

\bibliographystyle{IEEEtran}
\bibliography{references}

\begin{thebibliography}{10}
\providecommand{\url}[1]{#1}
\csname url@samestyle\endcsname
\providecommand{\newblock}{\relax}
\providecommand{\bibinfo}[2]{#2}
\providecommand{\BIBentrySTDinterwordspacing}{\spaceskip=0pt\relax}
\providecommand{\BIBentryALTinterwordstretchfactor}{4}
\providecommand{\BIBentryALTinterwordspacing}{\spaceskip=\fontdimen2\font plus
\BIBentryALTinterwordstretchfactor\fontdimen3\font minus \fontdimen4\font\relax}
\providecommand{\BIBforeignlanguage}[2]{{%
\expandafter\ifx\csname l@#1\endcsname\relax
\typeout{** WARNING: IEEEtran.bst: No hyphenation pattern has been}%
\typeout{** loaded for the language `#1'. Using the pattern for}%
\typeout{** the default language instead.}%
\else
\language=\csname l@#1\endcsname
\fi
#2}}
\providecommand{\BIBdecl}{\relax}
\BIBdecl

\bibitem{bohg_interactive_2017}
J.~Bohg, K.~Hausman, B.~Sankaran, O.~Brock, D.~Kragic, S.~Schaal, and G.~S. Sukhatme, ``Interactive {Perception}: {Leveraging} {Action} in {Perception} and {Perception} in {Action},'' \emph{IEEE Transactions on Robotics}, vol.~33, no.~6, pp. 1273--1291, Dec. 2017.

\bibitem{doshi-velez_hidden_2013}
F.~Doshi-Velez and G.~Konidaris, ``Hidden parameter markov decision processes: {A} semiparametric regression approach for discovering latent task parametrizations,'' in \emph{Proceedings of the {International} {Joint} {Conferences} on {Artificial} {Intelligence}}, 2016.

\bibitem{zhang_application_2012a}
L.~Zhang and J.~C. Trinkle, ``The application of particle filtering to grasping acquisition with visual occlusion and tactile sensing,'' in \emph{Proceedings of the {{IEEE International Conference}} on {{Robotics}} and {{Automation}}}, May 2012, pp. 3805--3812.

\bibitem{kaelbling_planning_1998}
\BIBentryALTinterwordspacing
L.~P. Kaelbling, M.~L. Littman, and A.~R. Cassandra, ``Planning and acting in partially observable stochastic domains,'' \emph{Artificial Intelligence}, vol. 101, no.~1, pp. 99--134, 1998. [Online]. Available: \url{https://www.sciencedirect.com/science/article/pii/S000437029800023X}
\BIBentrySTDinterwordspacing

\bibitem{sunberg_online_2018}
\BIBentryALTinterwordspacing
Z.~Sunberg and M.~Kochenderfer, ``\BIBforeignlanguage{en}{Online {Algorithms} for {POMDPs} with {Continuous} {State}, {Action}, and {Observation} {Spaces}},'' \emph{\BIBforeignlanguage{en}{Proceedings of the International Conference on Automated Planning and Scheduling}}, vol.~28, pp. 259--263, Jun. 2018. [Online]. Available: \url{https://ojs.aaai.org/index.php/ICAPS/article/view/13882}
\BIBentrySTDinterwordspacing

\bibitem{kim_attentive_2018}
\BIBentryALTinterwordspacing
H.~Kim, A.~Mnih, J.~Schwarz, M.~Garnelo, A.~Eslami, D.~Rosenbaum, O.~Vinyals, and Y.~W. Teh, ``\BIBforeignlanguage{en}{Attentive {Neural} {Processes}},'' in \emph{\BIBforeignlanguage{en}{NeurIPS Workshop on Bayesian Deep Learning}}, Sep. 2018. [Online]. Available: \url{https://openreview.net/forum?id=SkE6PjC9KX}
\BIBentrySTDinterwordspacing

\bibitem{killian_robust_2017}
T.~W. Killian, S.~Daulton, G.~Konidaris, and F.~Doshi-Velez, ``Robust and {Efficient} {Transfer} {Learning} with {Hidden} {Parameter} {Markov} {Decision} {Processes},'' in \emph{Proceedings of the 31st {Conference} on {Neural} {Information} {Processing} {Systems}}, 2017.

\bibitem{saemundsson_meta_2018}
S.~Sæmundsson, K.~Hofmann, and M.~P. Deisenroth, ``Meta reinforcement learning with latent variable gaussian processes,'' in \emph{{Uncertainty} in {Artificial} {Intelligence}}, 2018.

\bibitem{belkhale_model-based_2021}
S.~Belkhale, R.~Li, G.~Kahn, R.~McAllister, R.~Calandra, and S.~Levine, ``Model-{Based} {Meta}-{Reinforcement} {Learning} for {Flight} with {Suspended} {Payloads},'' \emph{IEEE Robotics and Automation Letters}, vol.~6, no.~2, 2021.

\bibitem{battaglia_simulation_2013}
P.~W. Battaglia, J.~B. Hamrick, and J.~B. Tenenbaum, ``Simulation as an engine of physical scene understanding,'' \emph{Proceedings of the National Academy of Sciences}, vol. 110, no.~45, 2013.

\bibitem{wu_galileo_2015}
J.~Wu, I.~Yildirim, J.~J. Lim, B.~Freeman, and J.~Tenenbaum, ``\BIBforeignlanguage{en}{Galileo: {Perceiving} {Physical} {Object} {Properties} by {Integrating} a {Physics} {Engine} with {Deep} {Learning}},'' in \emph{\BIBforeignlanguage{en}{Advances in Neural Information Processing Systems}}, 2015, p.~9.

\bibitem{wu_physics_2016}
J.~Wu, J.~Lim, H.~Zhang, J.~Tenenbaum, and W.~Freeman, ``Physics 101: {Learning} {Physical} {Object} {Properties} from {Unlabeled} {Videos},'' in \emph{Procedings of the {British} {Machine} {Vision} {Conference} 2016}, 2016.

\bibitem{zhang_application_2012}
L.~Zhang and J.~C. Trinkle, ``The application of particle filtering to grasping acquisition with visual occlusion and tactile sensing,'' in \emph{Proceedings of the {IEEE} {International} {Conference} on {Robotics} and {Automation}}, 2012.

\bibitem{lu_extracting_2019}
P.~Y. Lu, S.~Kim, and M.~Solja{\v{c}}i{\'c}, ``Extracting interpretable physical parameters from spatiotemporal systems using unsupervised learning,'' \emph{Physical Review X}, vol.~10, no.~3, p. 031056, 2020.

\bibitem{veerapaneni_entity_2019}
R.~Veerapaneni, J.~D. Co-Reyes, M.~Chang, M.~Janner, C.~Finn, J.~Wu, J.~Tenenbaum, and S.~Levine, ``Entity {Abstraction} in {Visual} {Model}-{Based} {Reinforcement} {Learning},'' in \emph{Proceedings of the Conference on {Robot} {Learning}}, 2019.

\bibitem{xu_densephysnet_2019}
Z.~Xu, J.~Wu, A.~Zeng, J.~B. Tenenbaum, and S.~Song, ``\BIBforeignlanguage{en}{{DensePhysNet}: {Learning} {Dense} {Physical} {Object} {Representations} via {Multi}-step {Dynamic} {Interactions}},'' in \emph{\BIBforeignlanguage{en}{Proceedings of {Robotics}: {Science} and {Systems}}}, 2019.

\bibitem{zheng_unsupervised_2018}
D.~Zheng, V.~Luo, J.~Wu, and J.~B. Tenenbaum, ``Unsupervised {Learning} of {Latent} {Physical} {Properties} {Using} {Perception}-{Prediction} {Networks},'' in \emph{{Uncertainty} in {Artificial} {Intelligence}}, 2018.

\bibitem{noseworthy_amortized_2024}
M.~Noseworthy, S.~Shaw, C.~C. Kessens, and N.~Roy, ``Amortized {{Inference}} for {{Efficient Grasp Model Adaptation}},'' \emph{International Conference on Robotics and Automation}, 2024.

\bibitem{danielczuk_exploratory_2021}
M.~Danielczuk, A.~Balakrishna, D.~Brown, and K.~Goldberg, ``Exploratory {Grasping}: {Asymptotically} {Optimal} {Algorithms} for {Grasping} {Challenging} {Polyhedral} {Objects},'' in \emph{Proceedings of the {Conference} on {Robot} {Learning}}, 2021.

\bibitem{fu_legs_2022}
L.~Fu, M.~Danielczuk, A.~Balakrishna, D.~S. Brown, J.~Ichnowski, E.~Solowjow, and K.~Goldberg, ``{LEGS}: {Learning} {Efficient} {Grasp} {Sets} for {Exploratory} {Grasping},'' in \emph{Proceedings of IEEE {International} {Conference} on {Robotics} and {Automation}}, 2022.

\bibitem{zhao_center_2018}
Z.~Zhao, X.~Li, C.~Lu, and Y.~Wang, ``Center of mass and friction coefficient exploration of unknown object for a robotic grasping manipulation,'' in \emph{Proceedings of the IEEE International Conference on Mechatronics and Automation}, 2018, pp. 2352--2357.

\bibitem{cehajic_estimating_2017}
D.~{\'C}ehaji{\'c}, S.~Hirche \emph{et~al.}, ``Estimating unknown object dynamics in human-robot manipulation tasks,'' in \emph{Proceedings of the IEEE International Conference on Robotics and Automation}, 2017, pp. 1730--1737.

\bibitem{herzog_learning_2014}
A.~Herzog, P.~Pastor, M.~Kalakrishnan, L.~Righetti, J.~Bohg, T.~Asfour, and S.~Schaal, ``\BIBforeignlanguage{en}{Learning of grasp selection based on shape-templates},'' \emph{\BIBforeignlanguage{en}{Autonomous Robots}}, vol.~36, no.~1, pp. 51--65, Jan. 2014.

\bibitem{dragiev_uncertainty_2013}
S.~Dragiev, M.~Toussaint, and M.~Gienger, ``Uncertainty aware grasping and tactile exploration,'' in \emph{Proceedings of the {IEEE} {International} {Conference} on {Robotics} and {Automation}}, May 2013.

\bibitem{gadre_act_2021}
S.~Y. Gadre, K.~Ehsani, and S.~Song, ``Act the part: Learning interaction strategies for articulated object part discovery,'' in \emph{Proceedings of the IEEE/CVF International Conference on Computer Vision}, 2021.

\bibitem{martin-martin_coupled_2022}
R.~Martín-Martín and O.~Brock, ``Coupled recursive estimation for online interactive perception of articulated objects,'' \emph{The International Journal of Robotics Research}, vol.~41, no.~8, pp. 741--777, Jul. 2022.

\bibitem{bauzaGPSUMGaussian2020}
M.~Bauza and A.~Rodriguez, ``{{GP-SUM}}. {{Gaussian Process Filtering}} of non-{{Gaussian Beliefs}},'' in \emph{Algorithmic {{Foundations}} of {{Robotics XIII}}}, M.~Morales, L.~Tapia, G.~{S{\'a}nchez-Ante}, and S.~Hutchinson, Eds.\hskip 1em plus 0.5em minus 0.4em\relax Cham: Springer International Publishing, 2020, pp. 508--525.

\end{thebibliography}

\end{document}